\documentclass[nohyperref]{article}
\usepackage{amsfonts}       % blackboard math symbols
\usepackage{nicefrac}       % compact symbols for 1/2, etc.
\usepackage{microtype}
\usepackage{subfigure}
\usepackage{booktabs}
\usepackage[colorlinks = true,
            linkcolor = black,
            urlcolor  = blue,
            citecolor = blue,
            anchorcolor = blue,
            pagebackref = true]{hyperref}
\usepackage{graphicx}
\usepackage{hyperref}
\usepackage{url}
\usepackage{amsthm}
\usepackage{amssymb}
\usepackage{amsmath}
\usepackage{xcolor}
\usepackage{mathrsfs}
\usepackage{bbm}
\usepackage{xcolor}
\usepackage{sidecap}
\usepackage{thm-restate}
\usepackage{enumerate}
\usepackage{float}
\usepackage{graphicx}
\usepackage{bbm}
\usepackage{caption}
\usepackage{algorithm}
\usepackage[noend]{algorithmic}
\usepackage{setspace}
\usepackage{authblk}
\usepackage{mathtools}
\usepackage{pythonhighlight}
\usepackage{wrapfig}
\usepackage{amsmath}
\mathtoolsset{showonlyrefs=true}

\sidecaptionvpos{figure}{c}

\newcommand{\E}{\mathop{\mathbb{E}}}
\newcommand{\bellman}{\mathcal{T}}

\newcommand{\reward}{r}
\newcommand{\transition}{\mathcal{P}}
\newcommand{\states}{\mathcal{S}}
\newcommand{\actions}{\mathcal{A}}

\newcommand{\aquadem}{AQuaDem}

\newcommand{\demos}{\mathcal{D}}

\usepackage[accepted]{icml2022}

\newboolean{arxiv}
\setboolean{arxiv}{true}

\usepackage[textsize=tiny]{todonotes}

\icmltitlerunning{Continuous Control with Action Quantization from Demonstrations}

\begin{document}

\twocolumn[
\icmltitle{Continuous Control with Action Quantization from Demonstrations}



\icmlsetsymbol{equal}{*}

\begin{icmlauthorlist}
\icmlauthor{Robert Dadashi}{equal,google}
\icmlauthor{L\'eonard Hussenot}{equal,google,lille}
\icmlauthor{Damien Vincent}{google}
\icmlauthor{Sertan Girgin}{google}\\
\icmlauthor{Anton Raichuk}{google}
\icmlauthor{Matthieu Geist}{google}
\icmlauthor{Olivier Pietquin}{google}
\end{icmlauthorlist}

\icmlaffiliation{google}{Google Research, Brain Team}
\icmlaffiliation{lille}{Univ. de Lille, CNRS, Inria Scool, UMR 9189 CRIStAL}

\icmlcorrespondingauthor{Robert Dadashi}{dadashi@google.com}

\icmlkeywords{Machine Learning, ICML, Reinforcement Learning, Imitation Learning, Discretization}

\vskip 0.3in
]



\printAffiliationsAndNotice{\icmlEqualContribution} % otherwise use the standard text.

\begin{abstract}
In this paper, we propose a novel Reinforcement Learning (RL) framework for problems with continuous action spaces: Action Quantization from Demonstrations (\aquadem{}). The proposed approach consists in learning a discretization of continuous action spaces from human demonstrations. This discretization returns a set of plausible actions (in light of the demonstrations) for each input state, thus capturing the priors of the demonstrator and their multimodal behavior. By discretizing the action space, any discrete action deep RL technique can be readily applied to the continuous control problem. Experiments show that the proposed approach outperforms state-of-the-art methods such as SAC in the RL setup, and GAIL in the Imitation Learning setup. We provide a website with interactive videos: \url{https://google-research.github.io/aquadem/} and make the code available: \url{https://github.com/google-research/google-research/tree/master/aquadem}.
\end{abstract}

\section{Introduction}
With several successes on highly challenging tasks including strategy games such as Go \citep{silver2016mastering}, StarCraft \citep{vinyals2019grandmaster} or Dota 2 \citep{berner2019dota} as well as robotic manipulation \citep{andrychowicz2020learning}, Reinforcement Learning (RL) holds a tremendous potential for solving sequential decision making problems. RL relies on Markov Decision Processes (MDP) \citep{puterman2014markov} as its cornerstone, a general framework under which vastly different problems can be casted.

There is a clear separation in the class of MDPs between the discrete action setup, where an agent faces a countable set of actions, and the continuous action setup, where an agent faces an uncountable set of actions. When the number of actions is finite and small, computing the maximum of the action-value function is straightforward (and implicitly defines a greedy policy). In the continuous action setup, the parametrized policy either directly optimizes the expected value function that is estimated through Monte Carlo rollouts \citep{williams1992simple}, which makes it demanding in interactions with the environment, or optimizes a parametrized state-action value function \citep{konda2000actor} hence introducing additional sources of approximations. 

Therefore, a workaround consists in turning a continuous control problem into a discrete one. The simplest approach is to naively (\textit{e.g.} uniformly) discretize the action space, an idea which dates back to the ``bang-bang'' controller \citep{bushaw1952differential}. However, such a discretization scheme suffers from the curse of dimensionality. Various methods have addressed this limitation by making the strong assumption of independence~\citep{tavakoli2018action,tang2020discretizing,andrychowicz2020learning} or of causal dependence~\citep{metz2017discrete, vinyals2019grandmaster, pmlr-v119-sakryukin20a, tavakoli2020learning} between the action dimensions which are typically complex and task-specific (e.g. autoregressive policies, pointer networks based architectures). 

In this work, we introduce a novel approach leveraging the prior of human demonstrations for reducing a continuous action space to a discrete set of meaningful actions. Demonstrations typically consist of transitions experienced by a human in the targeted environment, performing the task at hand or interacting without any specific goal \textit{i.e.} play. The side effect of using demonstrations to discretize the action space is to filter out useless/hazardous actions, thus focusing the search on relevant actions and possibly facilitating exploration. Besides, using a set of actions rather than a single one (Behavioral Cloning) enables to capture the multimodality of behaviors in the demonstrations.

\begin{figure*}[t]
\centering
\includegraphics[width=\linewidth]{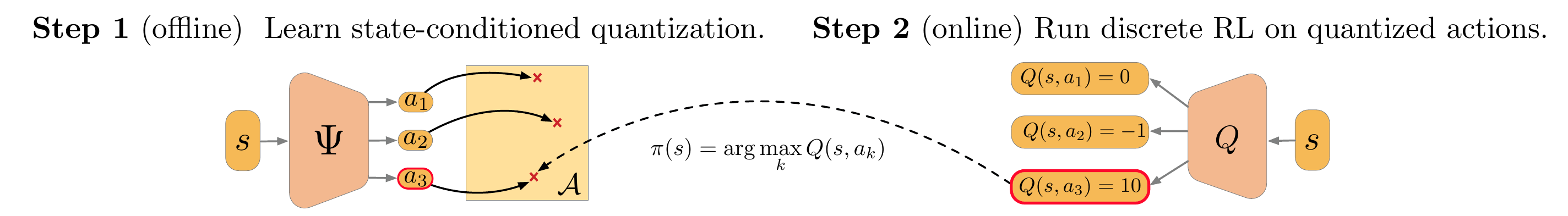}
\caption{Visualization of the \aquadem{} framework (offline) with a downstream algorithm (online).}
\label{fig:networks}
\end{figure*}

We thus propose  \underline{A}ction \underline{\smash{Qua}}ntization from \underline{Dem}onstrations, or \aquadem{}, a novel paradigm where we learn a state dependent discretization of a continuous action space using demonstrations, enabling the use of discrete-action deep RL methods. We formalize this paradigm, provide a neural implementation and analyze it through visualizations in simple grid worlds. We empirically evaluate this discretization strategy on three downstream task setups: Reinforcement Learning with demonstrations, Reinforcement Learning with play data (demonstrations of a human playing in an environment but not solving any specific task), and Imitation Learning. We test the resulting methods on challenging manipulation tasks and show that they outperform state-of-the-art continuous control methods both in terms of sample-efficiency and performance on every setup.

\section{Preliminaries}
\paragraph{Markov Decision Process.} We model the sequential decision making problem as a Markov Decision Process (MDP)~\citep{puterman2014markov, sutton2018reinforcement}. An MDP is a tuple $(\states, \actions, \transition, \reward, \gamma, \rho_0)$, where $\states$ is the state space, $\actions$ is the action space, $\transition$ is the transition kernel, $\reward$ is the expected reward function, $\gamma$ the discount factor and $\rho_0$ the initial state distribution. Throughout the paper, we distinguish discrete action spaces, which simply amount to a set $\{1,\dots,K\}$, from continuous action spaces which consist in an interval of $\mathbb{R}^d$ where $d$ is the dimensionality of the action space. A stationary policy $\pi$ is a mapping from states to distributions over actions. The value function $V^\pi$ of a policy $\pi$ is defined as the expected discounted cumulative reward from starting in a particular state and acting according to $\pi$:
$V^\pi(s) = \E \big[\sum^{\infty}_{t=0}  \gamma^t r(s_t, a_t) |s_0=s, a_t\sim\pi(s_t), s_{t+1}\sim\transition(s_t, a_t))\big]$. An optimal policy $\pi^*$ maximizes the value function $V^{\pi^*}$ for all states. The action-value function $Q^\pi$ is defined as the expected discounted cumulative reward from starting in a particular state, taking an action and then acting according to $\pi$:
$Q^\pi(s, a) = r(s, a) + \gamma \E \big[V^\pi(s')| s' \sim \transition(s, a) \big]$.
\paragraph{Value-based RL.} The~\citet{bellman} operator $\bellman$  connects an action-value function $Q$ for the state-action pair $(s,a)$ to the action-value function in the subsequent state $s'$: $\bellman^\pi (Q)(s, a) := r(s, a) + \gamma \E\big[ Q(s', a') | a' \sim \pi(s), s' \sim \transition(s, a) \big]$. Value Iteration (VI) \citep{bertsekas2000dynamic} is the basis for methods using the Bellman equation to derive algorithms estimating the optimal policy $\pi^*$. The prototypical example is the $Q$-learning algorithm \citep{watkins1992q}, which is the basis of \textit{e.g.} DQN \citep{mnih2015human}, and consists in the repeated application of a stochastic approximation of the Bellman operator $Q(s,a) := r(s,a) + \gamma \max_{a'} Q(s', a')$, where $(s, a, s')$ is a transition sampled from the MDP. The $Q$-learning algorithm exemplifies two desirable traits of VI-inspired methods in discrete action spaces that are \textbf{1)} bootstrapping: the current $Q$-value estimate at the next state $s'$ is used to compute a finer estimate of the $Q$-value at state $s$, and \textbf{2)} the exact derivation of the maximum $Q$-value at a given state. For continuous action spaces, state-of-the-art methods \citep{haarnoja2018soft,fujimoto2018addressing} are also fundamentally close to a VI scheme, as they rely on Bellman consistency, with the difference being that the argument maximizing the $Q$-value, in other words the parametrized policy, is approximate.

\paragraph{Demonstration data.} Additional data consisting of transitions from an agent may be available. These demonstrations may contain the reward information or not. In the context of Imitation Learning \citep{pomerleau1991efficient,ng1999policy,ng2000algorithms,ziebart2008maximum}, the assumption is that the agent generating the demonstration data is near-optimal and that rewards are not provided. The objective is then to match the distribution of the agent with the one of the expert. In the context of Reinforcement Learning with demonstrations (RLfD) \citep{hester2018deep,vecerik2017leveraging}, demonstration rewards are provided. They are typically used in the form of auxiliary objectives together with a standard learning agent whose goal is to maximize the environment reward. In the context of Reinforcement Learning with play~\citep{lynch2020learning}, demonstration rewards are not provided as play data is typically not task-specific.

Demonstration data can come from various sources, although a common assumption is that it is generated by a single, unimodal Markovian policy. However, most of available data comes from agents that do not fulfill this condition. In particular, for human data, and even more so when coming from several individuals, the behavior generating the episodes may not be unimodal nor Markovian.

\section{Method}\label{sec:method}
In this section, we introduce the \aquadem{} framework and a practical neural network implementation together with an accompanying objective function. We provide a series of visualizations to study the candidate actions learned with \aquadem{} in gridworld experiments.

\subsection{AQuaDem: Action Quantization from Demonstrations}
Our objective is to reduce a continuous control problem to a discrete action one, on which we can apply discrete-action RL methods. Using demonstrations, we wish to assign to each state $s \in \states$ a set of $K$ candidate actions from $\actions$. The resulting action space is therefore a discrete finite set of $K$ state-conditioned vectors. In a given state $s \in \states$, picking action $k \in \{1,\dots,K\}$ stands for picking the  $k^\text{th}$ candidate action for that particular state. The \aquadem{} framework refers to the discretization of the action space, and the resulting discrete action algorithms used with \aquadem{} on continuous control tasks are detailed in Section~\ref{sec:experiments}. We propose to 
learn the discrete action space through a modified version of the Behavioral Cloning (BC) \citep{pomerleau1991efficient} reconstruction loss that captures the multimodality of demonstrations. Indeed the typical  BC implementation consists in building a deterministic mapping between states and actions $\Phi: \states \mapsto \actions$. 
But in practice, and in particular when the demonstrator is human, the demonstrator can take multiple actions in a given state (we say that its behavior is \textit{multimodal}) which are all relevant candidates for \aquadem{}. We thus learn a mapping $\Psi:\states \mapsto \actions^K$ from states to a set of $K$ candidate actions and optimize a reconstruction loss based on a soft minimum between the candidate actions and the demonstrated action.

Suppose we have a dataset of expert demonstrations $\demos = \{(s_i, a_i)\}_{1:n}$. In the continuous action setting, the vanilla BC approach consists in finding a parametrized function $f_\Phi$ that minimizes the reconstruction error between predicted actions and actions in the dataset $\demos$. To ease notations, we will conflate the function $f_\Phi$ with its parameters $\Phi$ and simply note it $\Phi: \states \mapsto \actions$. The objective is thus to minimize: $\min_\Phi \E_{s, a \sim \mathcal{D}} \|\Phi(s) - a \|^2$. Instead, we propose to learn a set of $K$ actions $\Psi_k(s)$ for each state by minimizing the following loss:
\begin{align}
\min_{\Psi} \E_{s, a \sim \mathcal{D}} \big[ - T \log \big( \sum^K_{k=1} \exp( \frac{-\|\Psi_k(s) - a \|^2}{T}) \big) \big] 
\label{eq:loss_1}
\end{align}
where the temperature $T$ is a hyperparameter. Equation \eqref{eq:loss_1} corresponds to minimizing a soft-minimum between the candidates actions ${\Psi_1(s), \dots, \Psi_K(s)}$ and the demonstrated action $a$. Note that with $K=1$, this is exactly the BC loss. The larger the temperature $T$ is, the more the loss imposes all candidate actions to be close to the demonstrated action $a$ thus reducing to the BC loss. The lower the temperature $T$ is, the more the loss only imposes a single candidate action to be close to the demonstrated action $a$. We provide empirical evidence of this phenomenon in Section~\ref{sec:viz} and provite a formal justification in Appendix~\ref{app:proof}. Equation~\eqref{eq:loss_1} is also interpretable in the context of Gaussian mixture models (see Appendix~\ref{sec:gmm}). The $\Psi$ function enables us to define a new MDP where the continuous action space is replaced by a discrete action space of size $K$ corresponding to the $K$ action candidates returned by $\Psi$ at each state.

\subsection{Visualization \label{sec:viz}}
\begin{figure}[h!]
\centering
\includegraphics[width=0.45\linewidth]{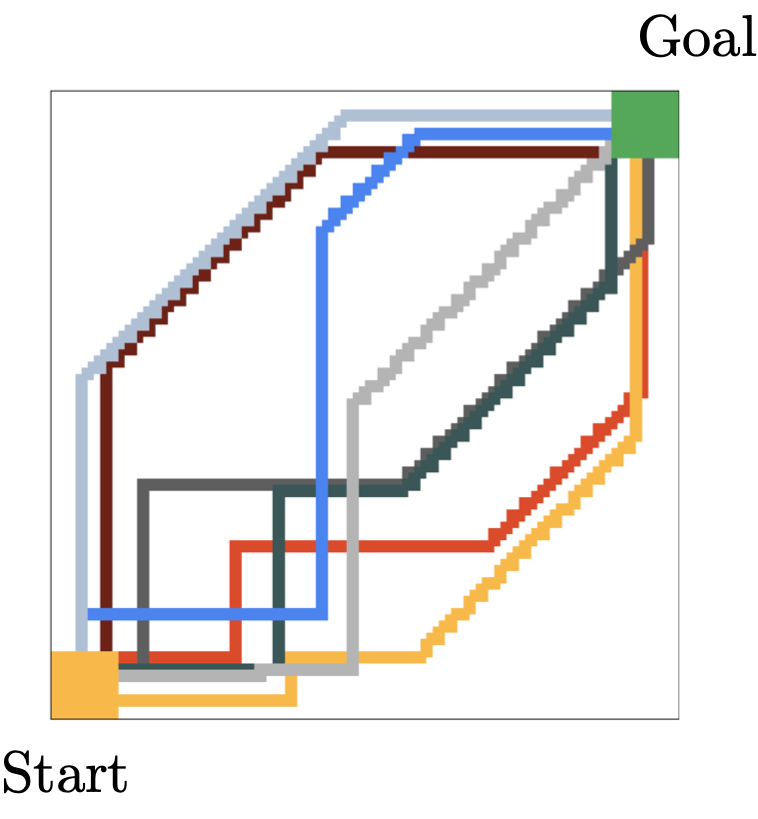}
\caption{Grid world environment with a start state (bottom left), and a goal state (top right). Actions are continuous, and give the direction in which the agent take a step with fixed length. The stochastic demonstrator moves either right or up in the bottom left of the environment then moves diagonally until reaching and edge and goes either up or right to the target. The demonstrations are represented in the different colors.}
\label{fig:grid_world}
\end{figure}

In this section, we analyze the actions learned through the \aquadem{} framework, in a toy grid world environment. We introduce a continuous action grid world with demonstrations in Figure~\ref{fig:grid_world}.

\begin{figure}[h!]
\centering
\includegraphics[width=\linewidth]{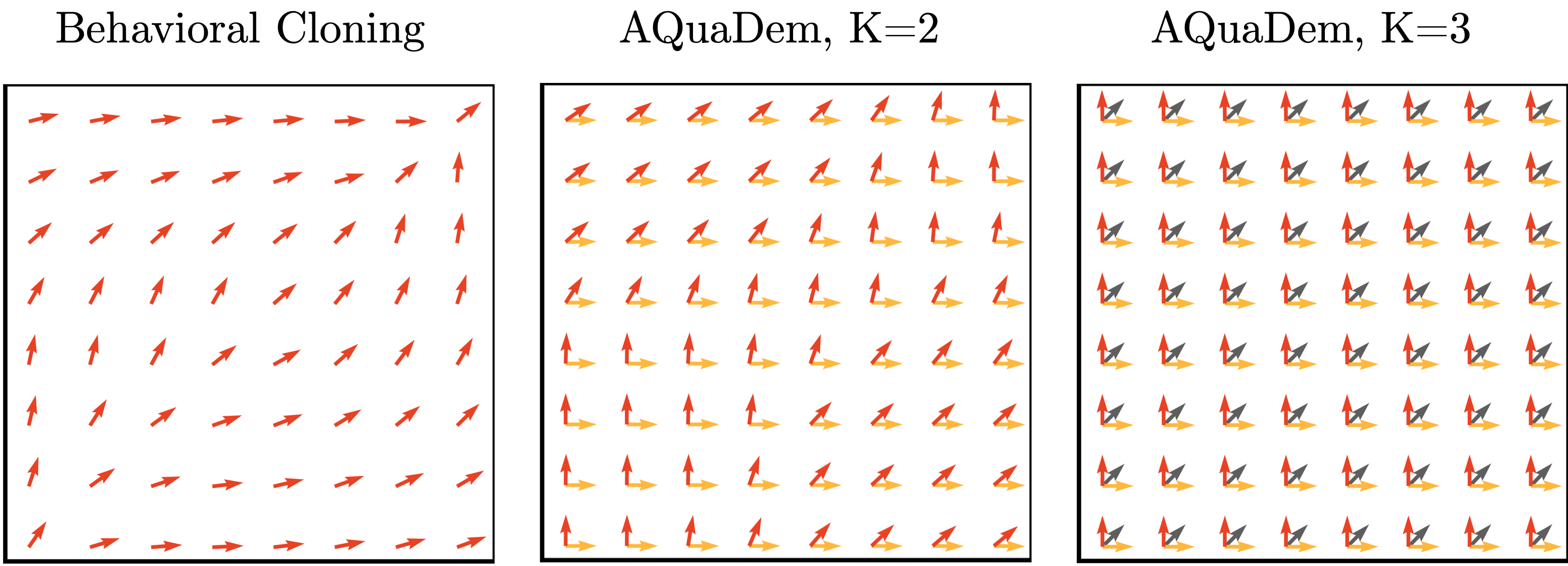}
\caption{Visualisation of the actions learned by BC and the candidate actions learned with \aquadem{} for $K=2$ and $K=3$ and $T=0.01$. Each color represents a head of the $\Psi$ network.}
\label{fig:actions}
\end{figure}

We define a neural network $\Psi$ and optimize its parameters by minimizing the objective function defined in Equation~\eqref{eq:loss_1} (implementation details can be found in Appendix~\ref{sec:gridworld_implem_details}). We display the resulting candidate actions across the state space in Figure~\ref{fig:actions}. As each color of the arrows depicts a single head of the $\Psi$ network, we observe that the action candidates are \textit{smooth}: action candidates slowly vary as the state vary, which prevents to have inconsistent action candidates in nearby states. Note that BC actions tend to be \textit{diagonal} even in the bottom left part of the action space, where the demonstrator only takes horizontal or vertical actions. On the contrary, the action candidates learned by \aquadem{} include the actions taken by the demonstrator conditioned on the states. Remark that in the case of $K=2$, the action \textit{right} is learned independently of the state position (middle plot in Figure~\ref{fig:actions}) although it is only executed in a subspace of the action space. In the case of $K=3$, actions are completely state-independent. In non-trivial tasks, the state dependence induced by the \aquadem{} framework is essential, as we show in the ablation study in Appendix~\ref{sec:ablations} and in the analysis of the actions learned in a more realistic setup in Appendix~\ref{sec:door_actions}.

\begin{figure}[h!]
\centering
\includegraphics[width=0.75\linewidth]{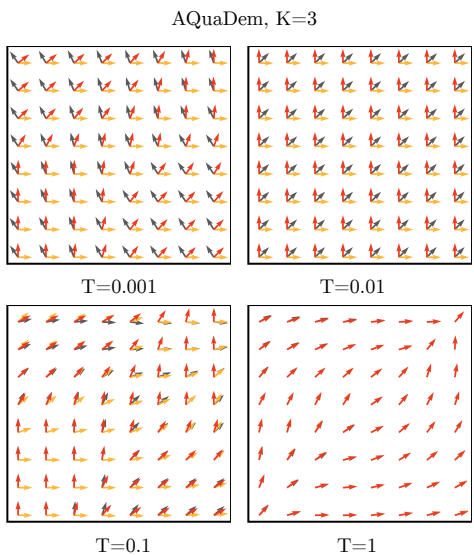}
\caption{Influence of the temperature on resulting candidate actions learned with \aquadem{}.}
\label{fig:temp_influence}
\end{figure}

\paragraph{Influence of the temperature.} The temperature controls the degree of smoothness of the soft-minimum defined in Equation~\eqref{eq:loss_1}. We show that with larger temperatures, the soft-minimum converges to the average which is well represented in Figure~\ref{fig:temp_influence} rightmost plot where the profile of \aquadem{}'s action candidates conflate with actions learned by BC. With lower temperatures, the actions taken by the demonstrator are recovered, but if the temperature is too low ($T=0.001$), some actions that are not taken by the demonstrator might appear as candidates (blue arrows in the leftmost figure). This occurs because the soft minimum converges to a hard minimum with lower temperatures meaning that as long as one candidate is close enough to the demonstrated action, the other candidates can be arbitrarily far off. In this work, we treat the temperature as a hyperparameter, although a natural direction for future work is to aggregate actions learned for different temperatures.

\subsection{Discussion}

\paragraph{On \textit{losing} the optimal policy.} In any form of discretization scheme, the resulting class of policies might not include the optimal policy of the original MDP. In the case of \aquadem{}, this mainly depends on the quality of the demonstrations. For standard continuous control methods, the parametrization of the policy also constrains the space of possible policies, potentially not including the optimal one. This is a lesser problem since policies tend to be represented with functions with universal approximation capabilities. Nevertheless, for most continuous control methods, the policy improvement step is approximate, while in the case of \aquadem{} it is exact, since it amounts to selecting the argmax of the $Q$-values.

\paragraph{On the multimodality of demonstrations.} The multimodality of demonstrations enables us to define multiple plausible actions for the agent to take in a given state, guided by the priors of the demonstrations. We argue that the assumption of multimodality of the demonstrator should actually be systematic \citep{mandlekar2021matters}. Indeed, the demonstrator can be \textit{e.g.} non-Markovian, optimizing for something different than a reward function like curiosity \citep{barto2013novelty}, or they can be in the process of learning how to interact with the environment. When demonstrations are gathered from multiple demonstrators, this naturally leads to multiple modalities in the demonstrations. And even in the case where the demonstrator is optimal, multiple actions might be equally good (\textit{e.g.} in navigation tasks). Finally, the demonstrator can interact with an environment without any task specific intent, which we refer to as \textit{play} \citep{lynch2020learning} and also induces a multimodal behavior.

\begin{figure*}[t]
\centering
\includegraphics[width=\linewidth]{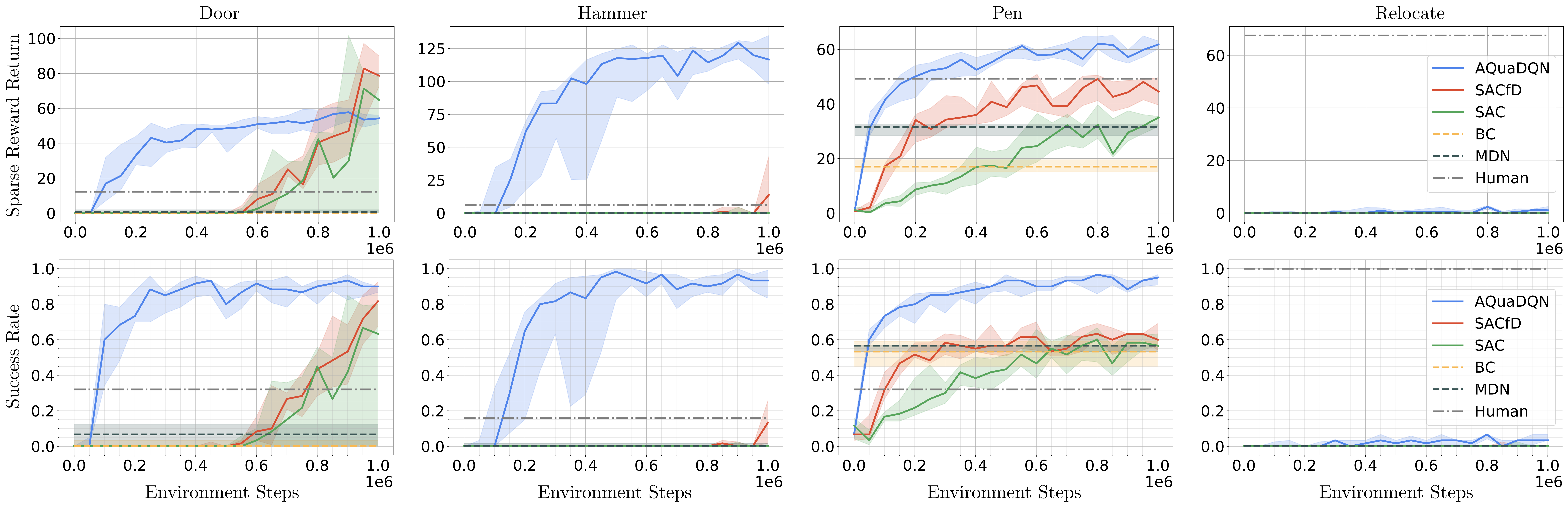}
\caption{Performance of AQuaDQN against SAC and SACfD. Agents are evaluated every $50$k environment steps over 30 episodes. We represent the median performance in terms of success rate (bottom) and returns (top) as well as the interquartile range over 10 seeds.}
\label{fig:lfd_results}
\end{figure*}

\section{Experiments} \label{sec:experiments}
In this section, we evaluate the \aquadem{} framework on three different downstream tasks setups: RL with demonstrations, RL with play data and Imitation Learning. For all experiments, we detail the networks architectures, hyperparameters search, and training procedures in the Appendix~\ifthenelse{\boolean{arxiv}}{and we provide videos of all the agents trained in the website.}{and we provide anonymized videos of the resulting agents and baselines; RL with demonstrations: \url{https://youtu.be/MyZzfA7RFnw}, Imitation Learning: \url{https://youtu.be/PxNIPx93slk} and RL with Play: \url{https://youtu.be/l6qbg54\_4eQ}.}  We also provide results for Offline RL in Appendix~\ref{sec:offline_rl}. 

\subsection{Reinforcement Learning with demonstrations}\label{sec:lfd}

\paragraph{Setup.} In the Reinforcement Learning with demonstrations setup (RLfD), the environment of interest comes with a reward function and demonstrations (which include the reward), and the goal is to learn a policy that maximizes the expected return. This setup is particularly interesting for sparse reward tasks, where the reward function is easy to define (say reaching a goal state) and where RL methods typically fail because the exploration problem is too hard. We consider the Adroit tasks \citep{rajeswaran2017learning} represented in Figure~\ref{fig:envs}, for which human demonstrations are available (25 episodes acquired using a virtual reality system). These environments come with a dense reward function that we replace with the following \textbf{sparse reward}: $1$ if the goal is achieved, $0$ otherwise.

\paragraph{Algorithm \& baselines.} The algorithm we propose is a two-fold training procedure: \textbf{1)} we learn a discretization of the action space in a fully offline fashion using the \aquadem{} framework from human demonstrations; \textbf{2)} we train a discrete action deep RL algorithm on top of this this discretization. We refer to this algorithm as AQuaDQN. The RL algorithm considered is Munchausen DQN \citep{vieillard2020munchausen} as it is the state of the art on the Atari benchmark \citep{bellemare2013arcade} (although we use the non-distributional version of it which simply amounts to DQN \citep{mnih2015human} with a regularization term). To make as much use of the demonstrations as possible, we maintain two replay buffers: one containing interactions with the environment, the other containing the demonstrations that we sample using a fixed ratio similarly to DQfD \citep{hester2018deep}, although we do not use the additional recipes of DQfD (multiple $n$-step evaluation of the bootstrapped estimate of $Q$, BC regularization term) for the sake of simplicity. When sampling demonstrations, the actions are discretized by taking the closest \aquadem{} action candidate (using the Euclidean norm). We consider SAC and SAC from demonstrations (SACfD) --a modified version of SAC where demonstrations are added to the replay buffer \citep{vecerik2017leveraging}-- as baselines against the proposed method. The implementation is from Acme~\citep{hoffman2020acme}. We do not include naive discretization baselines here, as the dimension of the action space is at least 24, which would lead to a $2^{24} \simeq 16$M actions with a binary discretization scheme, which is prohibitive without additional assumptions on the structure of the action-value function.
\begin{figure}[h!]
\centering
\includegraphics[width=0.65\linewidth]{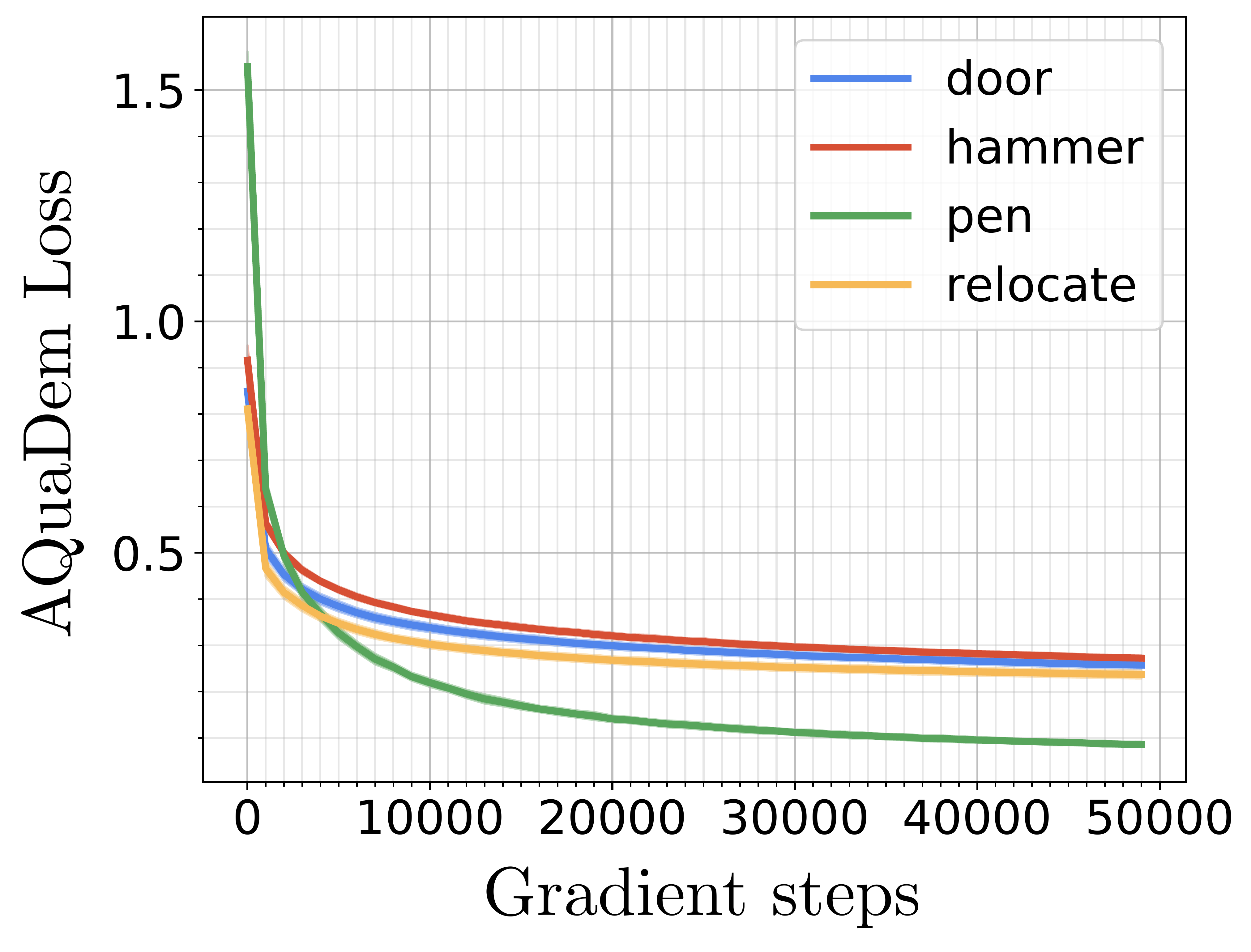}
\caption{AQuaDem discretization loss.}
\label{fig:aquadem_loss}
\end{figure}

\paragraph{Evaluation \& results.} We train the different methods on $1$M environment interactions on $10$ seeds for the chosen hyperparameters (a single set of hyperameters for all tasks) and evaluate the agents every $50$k environment interactions (without exploration noise) on $30$ episodes. An episode is considered a success if the goal is achieved during the episode. The \aquadem{} discretization is trained offline using $50$k gradient steps on batches of size $256$. The number of actions considered were $10,15,20$ and we found $10$ to be performing the best. Figure~\ref{fig:aquadem_loss} shows the \aquadem{} loss through the training procedure of the discretization step, and the Figure~\ref{fig:lfd_results} shows the returns of the trained agents as well as their success rate. On Door, Pen, and Hammer, the AQuaDQN agent reaches high success rate, largely outperforming SACfD in terms of success and sample efficiency.

\begin{figure}
\centering
\includegraphics[width=0.9\linewidth]{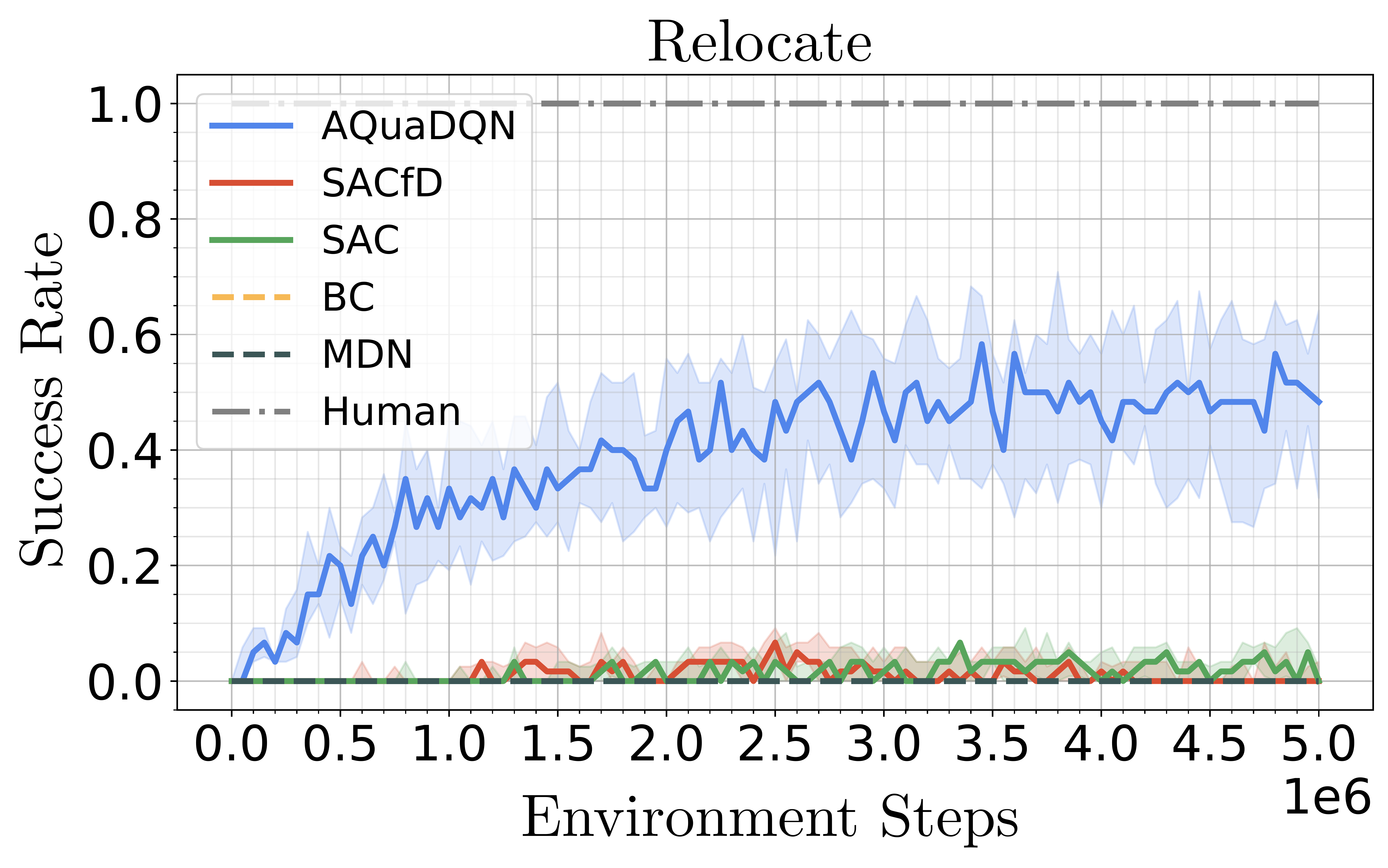}
\caption{Performance of AQuaDQN against SAC and SACfD baselines when all are tuned on the Relocate environment. We represent the median performance in terms of success rate as well as the interquartile range over 10 seeds.}
\label{fig:lfd_relocate}
\end{figure}

On Relocate, all methods reach poor results (although AQuaDQN slightly outperforms the baselines). The task requires a larger degree of generalisation than the other three since the goal state and the initial ball position are changing at each episode. We show in Figure~\ref{fig:lfd_relocate} that when tuned uniquely on the Relocate environment and with more environment interactions, AQuaDQN manages to reach a 50\% success rate where other methods still fail. Notice that on the Door environment, the SAC and SACfD agents outperform the AQuaDQN agent in terms of final return (but not in term of success rate). The behavior of these agents are however different from the demonstrator since they consist in slapping the handle and abruptly pulling it back. We provide videos of all resulting agents with one episode for each seed which is not cherry picked to demonstrate that AQuaDQN consistently learns a behavior that is qualitatively closer to the demonstrator (\url{https://youtu.be/MyZzfA7RFnw}).

\begin{figure*}[t]
\centering
\includegraphics[width=\linewidth]{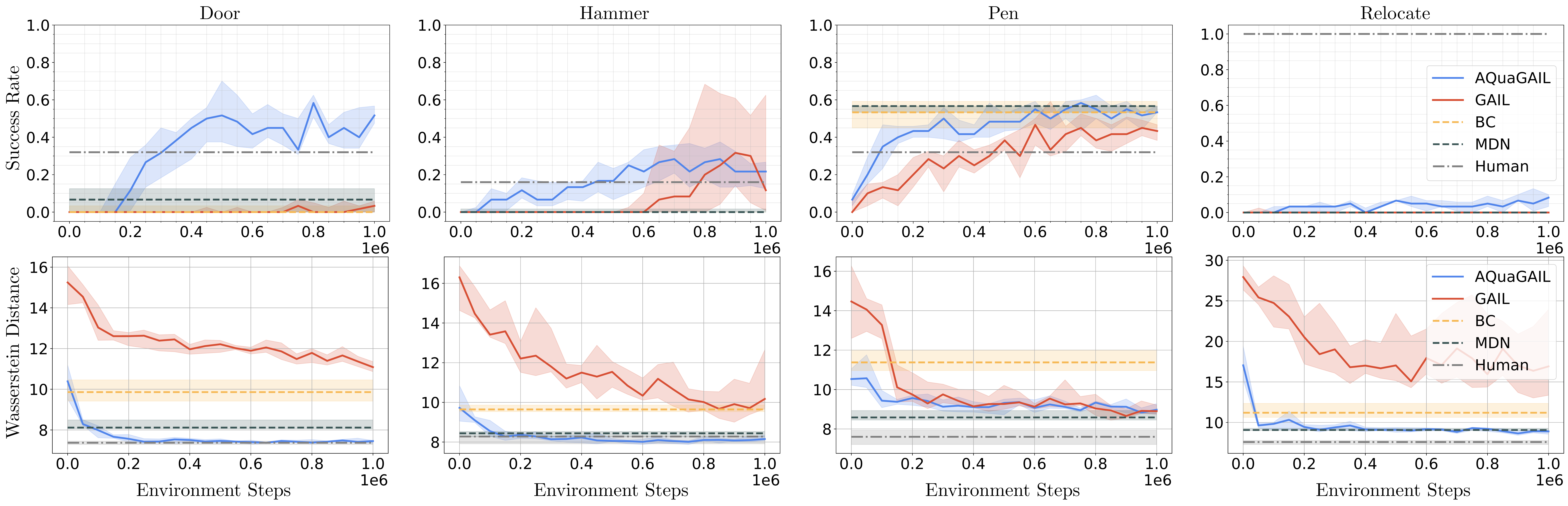}
\caption{Performance of AQuaGAIL against GAIL, BC and MDN baselines. Agents are evaluated every $50$k environment steps over 30 episodes. We represent the median success rate (top) on the task as well as the Wasserstein distance (bottom) of the agent's state distribution to the expert's state distribution as well as the interquartile range over 10 seeds.}
\label{fig:il_results}
\end{figure*}

\subsection{Imitation Learning \label{sec:il}}

\paragraph{Setup.} In Imitation Learning, the task is not specified by the reward function but by the demonstrations themselves. The goal is to mimic the demonstrated behavior. There is no reward function and the notion of success is ill-defined~\citep{hussenot2021hyperparameter}. A number of existing works \citep{ho2016generative,ghasemipour2019divergence,dadashi2020primal} cast the problem into matching the state distributions of the agent and of the expert.
Imitation Learning is of particular interest when designing a satisfying reward function --one that would lead the desired behavior to be the only optimal policy-- is harder than directly demonstrating this behavior. In this setup, there is no reward provided, not in the environment interactions nor in the demonstrations. We again consider the Adroit environments and the human demonstrations which consist of 25 episodes acquired via a virtual reality system.

\paragraph{Algorithm \& baselines.}
Again, the algorithm we propose has two stages. \textbf{1)} We learn --fully offline-- a discretization of the action space using \aquadem{}. \textbf{2)} We train a discrete action version of the GAIL algorithm~\citep{ho2016generative} in the discretized environment. More precisely, we interleave the training of a discriminator between demonstrations and agent experiences, and the training of a Munchausen DQN agent that maximizes the confusion of this discriminator. The Munchausen DQN takes one of the candidates actions given by \aquadem{}. We call this algorithm AQuaGAIL.
As a baseline, we consider the GAIL algorithm with a SAC~\citep{haarnoja2018soft} agent directly maximizing the confusion of the discriminator. This results in a very similar algorithm as the one proposed by \citet{kostrikov2018discriminator}. We also include the results of BC~\citep{pomerleau1991efficient} as well as BC with multiple  Mixture Density Networks (MDN) \citep{bishop1994mixture}.

\begin{figure*}[t]
\centering
 \includegraphics[width=\linewidth]{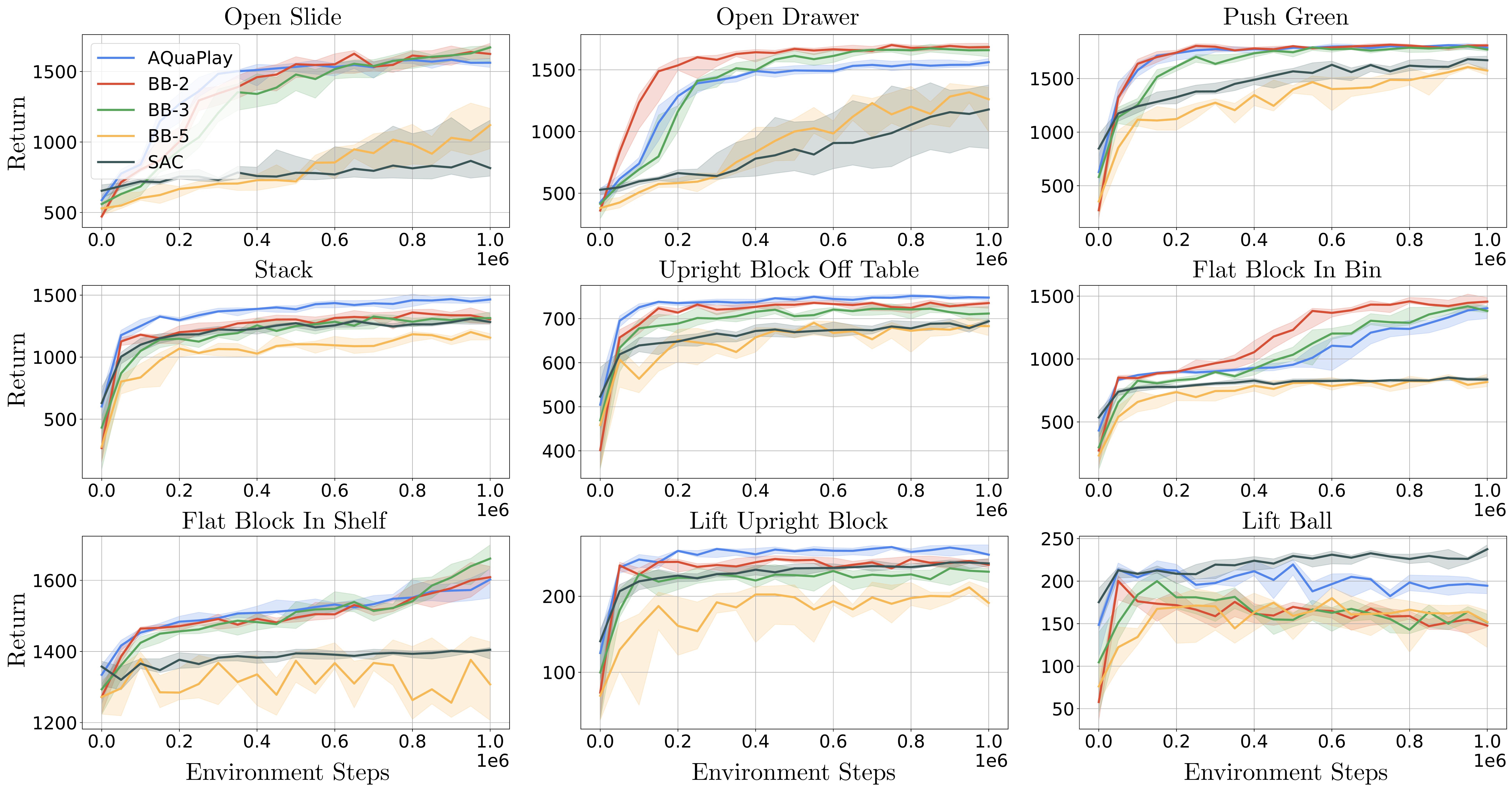}
\caption{Performance of AQuaPlay against SAC and ``bang-bang'' baselines. Agents are evaluated every $50$k environment steps over 30 episodes. We represent the median return as well as the interquartile range over 10 seeds.}
\label{fig:lfp_results}
\end{figure*}

\paragraph{Evaluation \& results.}
We train AQuaGAIL and GAIL for 1M environment interactions on 10 seeds for the selected hyperparameters (a single set for all tasks). BC and MDN are trained for 60k gradient steps with batch size $256$. We evaluate the agents every $50$k environment steps during training (without exploration noise) on 30 episodes. The \aquadem{} discretization is trained offline using $50$k gradient steps on batches of size $256$. The results are provided in Figure~\ref{fig:il_results}. Evaluating imitation learning algorithms has to be done carefully as the goal to ``mimic a behavior'' is ill-defined. Here, we provide the results according to two metrics. On top, the success rate is defined in Section~\ref{sec:lfd}. Notice that the human demonstrations do not have a success score of 1 on every task. We see that, except for Relocate, which is a hard task to solve with only 25 human demonstrations due to the necessity to generalize to new positions of the ball and the target, AQuaGAIL solves the tasks as successfully as the humans, outperforming GAIL, BC and MDN. Notice that our results corroborate previous work~\citep{orsini2021matters} that showed poor performance of GAIL on human demonstrations after 1M steps. The second metric we provide, on the bottom, is the Wasserstein distance between the state distribution of the demonstrations and the one of the agent. We compute it using the POT library~\citep{flamary2021pot} and use the Sinkhorn distance, a regularized version of the Wasserstein distance, as it is faster to compute. The ``human'' Wasserstein distance score is computed by randomly taking 5 episodes out of the 25 human demonstrations and compute the Wasserstein distance to the remaining 20. We repeat this procedure 100 times and plot the median (and the interquartile range) of the obtained values. Remark that AQuaGAIL is able to get much closer behavior to the human than BC, GAIL and MDN on all four environments in terms of Wasserstein distance. This supports that \aquadem{} leads to policies much closer to the demonstrator. We provide videos of the trained agents as an additional qualitative empirical evidence to support this claim (\url{https://youtu.be/PxNIPx93slk}).

\subsection{Reinforcement Learning with play data}

\paragraph{Setup.} The Reinforcement Learning with play data is an under-explored yet natural setup \citep{lynch2019play}. In this setup, the environment of interest has multiple tasks, a shared observation and action space for each task, and a reward function specific to each of the tasks. We also assume that we have access to \textit{play data}, introduced by \citet{lynch2020learning}, which consists in episodes from a human demonstrator interacting with an environment with the sole intention to play with it. The goal is to learn an optimal policy for each of the tasks. We consider the Robodesk tasks \citep{kannan2021robodesk} shown in Figure~\ref{fig:envs}, for which we acquired play data. We expand on the environment as well as the data collection procedure in the Appendix~\ref{sec:robodesk}.

\paragraph{Algorithm \& baselines.} Similarly to the RLfD setting, we propose a two-fold training procedure: \textbf{1)} we learn a discretization of the action space in a fully offline fashion using the \aquadem{} framework on the play data, \textbf{2)} we train a discrete action deep RL algorithm using this discretization on each tasks. We refer to this algorithm as AQuaPlay. Unlike the RLfD setting, the demonstrations do not include any task specific reward nor goal labels meaning that we cannot incorporate the demonstration episodes in the replay buffer nor use some form of goal-conditioned BC. We use SAC as a baseline, which is trained to optimize task specific rewards. Since the action space dimensionality is fairly low (5-dimensional), we can include naive uniform discretization baselines that we refer to as ``bang-bang''
~\citep{bushaw1952differential}. The original ``bang-bang'' controller (BB-2) is based on the extrema of the action space, we also provide a uniform discretization scheme based on 3 and 5 bins per action dimension, that we refer to as BB-3 and BB-5 respectively.

\paragraph{Evaluation \& results.} We train the different methods on $1$M environment interactions on $10$ seeds for the chosen hyperparameters (a single set of hyperameters for all tasks) and evaluate the agents every $50$k environment interactions (without exploration noise) on $30$ episodes. The \aquadem{} discretization is trained offline on play data using $50$k gradient steps on batches of size $256$. The number of actions considered were $10,20,30,40$ and we found $30$ to be performing the best. It is interesting to notice that it is higher than for the previous setups. It aligns with the intuition that with play data, several behaviors needs to be modelled. The results are provided in Figure~\ref{fig:lfp_results}. The AQuaPlay agent consistently outperforms SAC in this setup. Interestingly, the performance of the BB agent decreases with the discretization granularity, well exemplifying the curse of dimensionality of the method. In fact, BB with a binary discretization (BB-2) is competitive with AQuaPlay, which validates that discrete action RL algorithms are well performing if the discrete actions are sufficient to solve the task. Note however that the Robodesk environment is a relatively low-dimensional action environment, making it possible to have BB as a baseline, which is not the case of \textit{e.g.} Adroit where the action space is high-dimensional.

\section{Related Work} \label{sec:related}
\paragraph{Continuous action discretization.} The discretization of continuous action spaces has been introduced in control problems by \citet{bushaw1952differential} with the ``bang-bang'' controller \citep{bellman1956bang}. This naive discretization is problematic in high-dimensional action spaces, as the number of actions grows exponentially with the action dimensionality. To mitigate this phenomenon, a possible strategy is to assume that action dimensions are independent \citep{tavakoli2018action,andrychowicz2020learning,vieillard2021implicitly, tang2020discretizing}, or to assume or learn a causal dependence between them \citep{metz2017discrete,tessler2019distributional, pmlr-v119-sakryukin20a, tavakoli2020learning}. The \aquadem{} framework circumvents the curse of dimensionality as the discretization is based on the demonstrations and hence is dependent on the multimodality of the actions picked by the demonstrator rather than the dimensionality of the action space. Recently, \citet{seyde2021bang} replaced the Gaussian parametrization of continuous control methods, including SAC, by a Bernoulli parametrization to sample extremal actions (still relying on a sampling-based optimisation due to the high dimensionality of the action space), and show that performance remains similar on the DM control benchmark \citep{tassa2020dmcontrol}. The \aquadem{} framework does not favor extremal actions (which could be suboptimal) but the actions selected by the demonstrator instead. Close to our setup is the case where the action space is both discrete and continuous \citep{neunert2020continuous} or the action space is discrete and large \citep{dulac2015deep}. Those setups are interesting directions for extending \aquadem{}.

\paragraph{Q-learning in continuous action spaces.} Policy-based methods consist in solving continuous or discrete MDPs based on maximizing the expected return over the parameters of a family of policies. If the return is estimated through Monte Carlo rollouts, this leads to algorithms that are typically sample-inefficient and difficult to train in high-dimensional action spaces \citep{williams1992simple,schulman2015trust,schulman2017proximal}. As a result, a number of policy-based methods inspired from the policy gradient theorem \citep{sutton2000policy}, aim at maximizing the return using an approximate version of the  $Q$-value thus making them more sample-efficient. One common architecture is to parameterize a  $Q$-value, which is estimated by enforcing Bellman consistency, and define a policy using an optimization procedure of the parametrized  $Q$-value. Typical strategies to solve the  $Q$-value maximization include enforcing the  $Q$-value to be concave \citep{gu2016continuous,amos2017input} making it easy to optimize through \textit{e.g.} gradient ascent, to use a black box optimization method \citep{kalashnikov2018qt,simmons2019q,lim2018actor}, to solve a mixed integer programming problem \citep{ryu2019caql}, or to follow a biased estimate of the policy gradient based on the approximate  $Q$-value \citep{konda2000actor,lillicrap2015continuous,haarnoja2018soft,fujimoto2018addressing}. Recently, \citet{asadi2021deep} proposed to use a network that outputs actions together with their associated $Q$-values, tuned for each of the tasks at hand, on low-dimensional action spaces. Note that maximizing the approximate $Q$-value is a key problem that does not appear in \textit{small} discrete action spaces as in the \aquadem{} framework.

\paragraph{Hierarchical Imitation Learning.} A number of approaches have explored the learning of \textit{primitives} or \textit{options} from demonstrations together with a high-level controller that is either learned from demonstrations \citep{kroemer2015towards,krishnan2017ddco,le2018hierarchical,ding2019goal,lynch2020learning}, or learned from interactions with the environment \citep{manschitz2015learning,kipf2019compile,shankar2019discovering}, or hand specified \citep{pastor2009learning,fox2019multi}. \aquadem{} can be loosely interpreted as a two-level procedure as well, where the primitives (action discretization step) are learned fully offline, however there is no concept of goal nor temporally extended actions.

\paragraph{Modeling multimodal demonstrations.} A number of works have modeled the demonstrator data using multimodal architectures. For example, \citet{chernova2007confidence,calinon2007incremental} introduce Gaussian mixture models in their modeling of the demonstrator data. More recently, \citet{rahmatizadeh2018virtual} use Mixture density networks together with a recurrent neural network to model the temporal correlation of actions as well as their multimodality. \citep{yu2018one} also uses Mixture density networks to meta-learn a policy from demonstrations for one-shot adaptation. Another recent line of works has considered the problem of modeling demonstrations using an energy-based model, which is well adapted for multimodalities \citep{jarrett2020strictly,florence2021implicit}. \citet{singh2020parrot} also exploit the demonstrations prior for downstream tasks by learning a prior using a state-conditioned action generative model coupled with a continuous action algorithm. This is different from \aquadem{} that exploits the demonstrations prior to learn a discrete action space in order to use discrete action RL algorithms. 

\section{Perspectives and Conclusion} \label{sec:perspectives}
With the \aquadem{} paradigm, we provide a simple yet powerful method that enables to use discrete-action deep RL methods on continuous control tasks using demonstrations, thus escaping the complexity or curse of dimensionality of existing discretization methods. We showed in three different setups that it provides substantial gains in sample efficiency and performance and that it leads to qualitatively better agents. There are a number of different research avenues opened by \aquadem{}. Other discrete action specific methods could be leveraged in a similar way in the context of continuous control: count-based exploration \citep{tang2017exploration} or planning \citep{browne2012survey}. Similarly a number of methods in Imitation Learning \citep{brantley2019disagreement,wang2019random} or in offline RL \citep{fujimoto2021minimalist,wu2019behavior} are evaluated on continuous control tasks and are based on Behavioral Cloning regularization which could be refined using the same type of multioutput architecture used in this work. 
Finally, as the gain of sample efficiency is clear in different experimental settings, we believe that the \aquadem{} framework could be an interesting avenue for learning controllers on physical systems. 

\bibliography{main}
\bibliographystyle{icml2022}

%%%%%%%%%%%%%%%%%%%%%%%%%%%%%%%%%%%%%%%%%%%%%%%%%%%%%%%%%%%%%%%%%%%%%%%%%%%%%%%
%%%%%%%%%%%%%%%%%%%%%%%%%%%%%%%%%%%%%%%%%%%%%%%%%%%%%%%%%%%%%%%%%%%%%%%%%%%%%%%
% APPENDIX
%%%%%%%%%%%%%%%%%%%%%%%%%%%%%%%%%%%%%%%%%%%%%%%%%%%%%%%%%%%%%%%%%%%%%%%%%%%%%%%
%%%%%%%%%%%%%%%%%%%%%%%%%%%%%%%%%%%%%%%%%%%%%%%%%%%%%%%%%%%%%%%%%%%%%%%%%%%%%%%
\newpage
\appendix
\onecolumn

\section{Connection to Gaussian mixture models \label{sec:gmm}}
The BC loss can be interpreted as a maximum likelihood objective under the assumption that the demonstrator data comes from a Gaussian distribution. Similarly to Mixture density networks \citep{bishop1994mixture}, we propose to replace the Gaussian distribution by a mixture of Gaussian distributions. Suppose we represent the probability density of an action conditioned on a state by a mixture of $K$ Gaussian kernels: $p(a|s) = \sum_{k=1}^K \alpha_k(s) d_k(a|s)$, where $\alpha_k(s)$ is the mixing coefficient (that can be interpreted as a state conditioned prior probability), and $d_k(a|s)$ is the conditional density of the target $a$. Now assuming that the $K$ kernels are centered on ${\Psi_k(s)}_{k=1:K}$ and have fixed covariance $\sigma^2 \mathbbm{1}$ where $\sigma$ is a hyperparameter, we can write the log-likelihood of the demonstrations data $\demos$ as:
\begin{align}
    \mathcal{LL(\demos)} &= \sum_{s, a \in \demos} \log(p(s)p(a|s)) = \sum_{s, a \in \demos} \log p(s) + \log\big( \sum_{k=1}^K \alpha_k(s) d_k(a|s) \big) \\
    &= \sum_{s, a \in \demos} \log p(s) + \log\big(C\sum_{k=1}^K \alpha_k(s) \exp\big(-\frac{\|\Psi_k(s) - a|^2}{\sigma^2} \big) \big).
\end{align}
Therefore minimizing the negative log likelihood reduces to minimizing:
\begin{align}
 \sum_{s, a \in \demos} - \log\big(\sum_{k=1}^K \alpha_k(s) \exp\big(-\frac{\|\Psi_k(s) - a\|^2}{\sigma^2} \big) \big).
\end{align}
We propose to use a uniform prior $\alpha_k(s) = \frac{1}{K}$ when learning the locations of the centroids, which leads exactly to Equation \eqref{eq:loss_1} where the variance $\sigma^2$ is the temperature $T$. Note that we initially learned the state conditioned prior $\alpha_k(s)$, but we found no empirical evidence that it may be used to improve the performance of the downstream algorithms defined in Section~\ref{sec:experiments}.

\section{Asymptotic Behavior of the AQuaDem Loss \label{app:proof}}
For lighter notations, we write $x_k = || \Psi_k(s) -a ||^2$ and $x=(x_1,...,x_K)$.
For a single state-action pair (the empirical expectation being not relevant for studying the effect of the temperature), the AQuaDem loss can be rewritten:
\begin{align}
J(\Psi) &= - T\log \sum_{k=1}^K \exp(-\frac{x_k}{T}) = -T\log ( \frac{1}{K}\sum_{k=1}^K \exp(-\frac{x_k}{T})) - \log K.
\end{align}
Let's define $f_T(x) = -T\log ( \frac{1}{K}\sum_{k=1}^K \exp(-\frac{x_k}{T}))$.
The function $f_T$ is the same as the loss up to a constant term that does not change the solution of the optimization problem. Therefore, we can study this function for the behavior of the loss with respect to the temperature.

Now, denoting  $x_m = \min_k x_k$, we'll first study the behavior for low temperature.
\begin{align}
f_T(x) &= T \log K - T \log ( \sum_{k=1}^K \exp(-\frac{x_k}{T}))\\
&= T \log K - T \log (\exp(\frac{-x_m}{T}) \sum_{k=1}^K \exp(-\frac{x_k-x_m}{T}))\\
&= T \log K + x_m - T \log(1 + \sum_{k=1, k \neq m}^K \exp(-\frac{x_k-x_m}{T})) \xrightarrow[T\to 0]{} x_m.
\end{align}
Therefore, when the temperature goes to zero, $f_T$ behaves as the minimum.

For large temperatures, we have, using Taylor expansions:
\begin{align}
f_T(x) &= - T \log (\sum_{k=1}^K \frac{1}{K} \exp(-\frac{x_k}{T})) =- T \log (\sum_{k=1}^K \frac{1}{K} (1-\frac{x_k}{T} + o(\frac{x_k}{T})))\\
&=- T \log (1- \frac{1}{K} \sum_{k=1}^K \frac{x_k}{T} +o(\frac{1}{T})) = \frac{T}{K} \sum_{k=1}^K \frac{x_k}{T} + O(\frac{1}{T}) \xrightarrow[T\to \infty]{} \frac{1}{K} \sum_{k=1}^K x_k.
\end{align}
So, when the temperature goes to infinite, $f_T$ behaves as the average.

\section{Action visualization in a high-dimensional environment\label{sec:door_actions}}

For the Door environment (see Figure~\ref{fig:envs}) we represent the actions candidates learned using the AQuaDem framework with videos that can be found in the \ifthenelse{\boolean{arxiv}}{the website.}{supplementary material in the folder \texttt{visualizations} (for the best hyperparameters in the RL with demonstrations setting see Appendix~\ref{sec:lfd-aquadqn}).} As the action space is of high dimensionality, we choose to represent each action dimension on the x-axis, and the value for each dimension on the y-axis. We connect the dots on the x-axis to facilitate the visualization through time. We replay a trajectory from the human demonstrations and show at each step the 10 actions proposed by the AQuaDem network, and the action actually taken by the human demonstrator. Each action candidate has a color consistent across time (meaning that the blue action always correspond to the same head of the $\Psi$ network). Interestingly, the video shows that actions are very state dependent (except some default $0$-action) and evolve smoothly through time.

\section{Ablation Study\label{sec:ablations}}
In this section, we provide two ablations of the AQuaDQN algorithm. The first ablation is to learn a fixed set of actions independently of the state (which reduces to $K$-means). The second ablation consists in using random actions rather than the actions learned by the AQuaDem framework (the actions are given by the AQuaDem network, randomly initialized and not trained). We use the same hyperparameters as the one selected for AQuaDQN. In each case, for a number of actions in \{5, 10, 25\}, the success rate of the agent is 0 for all tasks throughout the training procedure. 

\section{Sanity Check Baselines}

\begin{figure}[h!]
\centering
\includegraphics[width=1.\linewidth]{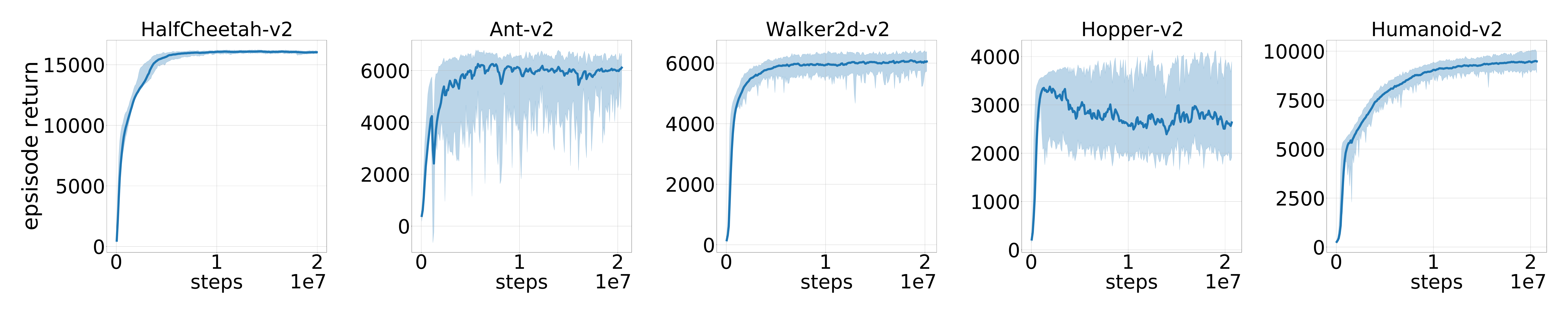}
\caption{SAC median and interquartile range on 10 seeds on the 5 Open Gym environments.}
\label{fig:mujoco_sac}
\end{figure}

We provide in Figure~\ref{fig:mujoco_sac} the results of the SAC implementation from Acme~\citep{hoffman2020acme} on the 5 classical OpenAI Gym environments, for 20M steps. The hyperparameters are the ones of the SAC paper \citep{haarnoja2018soft2} where the adaptive temperature was introduced. The results are consistent with the original paper and the larger relevant literature.

\section{Offline Reinforcement Learning \label{sec:offline_rl}}

In this section, we lead the analysis of the \aquadem{} framework in the context of Offline Reinforcement Learning \citep{levine2020offline}. In the following, we no longer assume that the agent can interact with the environment, and learn a policy from a fixed set of transitions $\demos$. We use the CQL \citep{kumar2020conservative} algorithm together with the \aquadem{} discretization. A simplistic variant of the CQL algorithm minimizes the following objective:
\begin{align}
    \min_{Q} \E_{s, a \sim \demos} \Big[ \alpha \big( \log \sum \exp(Q(s, .)) - Q(s, a) \big) + \frac{1}{2} (Q - \bellman^* Q)^2(s, a) \Big]. \label{eq:cql_loss}
\end{align}
The CQL loss defined in Equation \eqref{eq:cql_loss} is natural for discrete action space, due to the logsumexp term, which is the regularizer of the Q-values that prevents out-of-distribution extrapolation). In continuous action spaces, the logsumexp term needs to be estimated. In the authors' implementation, the sampling logic is intricate has it relies on 3 different distributions. We evaluate the resulting algorithm on the D4RL locomotion tasks and provide performance against state-of-the-art offline RL algorithms. We use the results reported by \citet{kostrikov2021offline}. We provide results in Table \ref{tab:aquacql} and show that AQuaCQL is competitive with considered baselines.

The architecture of the \aquadem{} network has a common 3 layers of size 256 with relu activation, and a subsequent hidden layer of size 256 with relu activation for each action. We do not use dropout regularization and trained the network for $5.10^6$ steps with batches of size 256. We use $\alpha=5$ in the CQL loss (Equation \eqref{eq:cql_loss}) and we use Munchausen DQN, with the same hyperparameters as AQuaDQN (Section~\ref{sec:lfd-aquadqn}) except for the discount factor ($\gamma = 0.995$) and train it for $10^6$ gradient steps.

\begin{table}[h!]
  \centering 
  \begin{tabular}{|l|l|l|l|l||l|}
    \toprule
    Environment & BC & TD3+BC & CQL & IQL & AQuaCQL \\
    \midrule
    halfcheetah-medium-v2 & 42.6 & \textbf{48.3} & 44.0 & \textbf{47.4} & 44.5 $\pm$ 2.5 \\
	hopper-medium-v2 & 52.9 & 59.3 & 58.5 & \textbf{66.3} & 58.5 $\pm$ 2.5 \\
	walker2d-medium-v2 & 75.3 & \textbf{83.7} & 72.5 & 78.3 & 82.1 $\pm$ 0.4\\
	halfcheetah-medium-replay-v2 & 36.6 & \textbf{44.6} & \textbf{45.5} & \textbf{44.2} & 40.5 $\pm$ 2.5 \\
	hopper-medium-replay-v2 & 18.1 & 60.9 & \textbf{95.0} & \textbf{94.7} & 90.3 $\pm$ 2.1 \\
	walker2d-medium-replay-v2 & 26.0 & \textbf{81.8} & 77.2 & 73.9 & 80.8 $\pm$ 1.43\\
	halfcheetah-medium-expert-v2 & 55.2 & \textbf{90.7} & \textbf{91.6} & 86.7 & \textbf{88.3 $\pm$ 3} \\
	hopper-medium-expert-v2 & 52.5 & 98.0 & \textbf{105.4} & 91.5 & 86.7 $\pm$ 6.9\\
	walker2d-medium-expert-v2 & \textbf{107.5} & \textbf{110.1} & \textbf{108.8}  &\textbf{109.6} & 108.1 $\pm$ 0.3 \\
    \bottomrule
    total & 466.7 & \textbf{677.4} & \textbf{698.5}  &\textbf{692.4} & \textbf{679.9 $\pm$ 22.}\\
    \bottomrule
  \end{tabular}
  \caption{Averaged normalized scores on MuJoCo locomotion tasks. The AQuaCQL results are averaged over 10 seeds and 10 evaluation episodes. \label{tab:aquacql}}
\end{table}

\section{Implementation}

\subsection{Grid World Visualizations \label{sec:gridworld_implem_details}}
We learn the discretization of the action space using the \aquadem{} framework. The architecture of the network is a common hidden layer of size $256$ with relu activation, and a subsequent hidden layer of size $256$ with relu activation for each action. We minimize the \aquadem{} loss using the Adam optimizer with the learning rate $0.0003$ and the dropout regularization rate $0.1$ on $20000$ gradient steps. 

\subsection{Environments \label{sec:robodesk}}

We considered the Adroit environments and the Robodesk environments, for which we described the observation space and the action space in Table \ref{table:envs}.

\begin{table}[h]
  \centering
  \begin{tabular}{lll}
    \toprule
    Environment & Observation Space & Action Space\\
    \midrule
    Door & 39 & 28\\
    Hammer & 46 & 26\\
    Pen & 45 & 24\\
    Relocate & 39 & 30\\
    Robodesk & 76 & 5\\
    \bottomrule
  \end{tabular}
  \caption{Environment description of the Adroit and Robodesk observation and action space.}
  \label{table:envs}
\end{table}

\begin{figure}[h!]
\centering
\includegraphics[width=1.\linewidth]{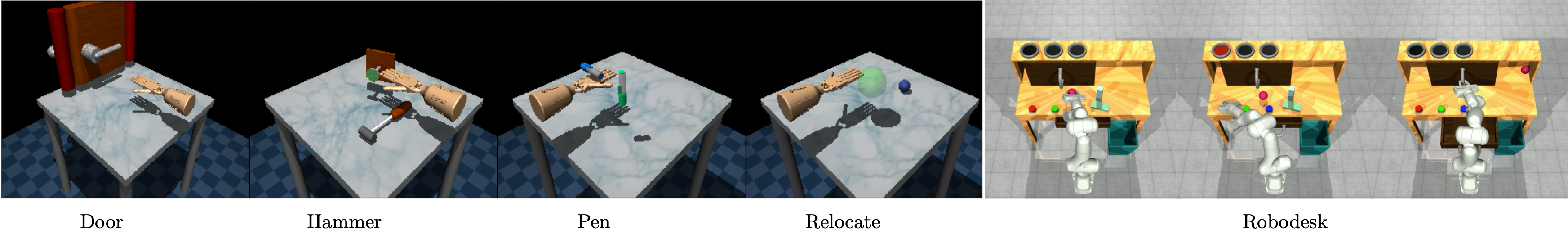}
\caption{Visualizations of the Adroit and Robodesk environments. }
\label{fig:envs}
\end{figure}

\paragraph{Adroit} The Adroit environments \citep{rajeswaran2017learning} consists in a shadow hand solving 4 tasks (Figure~\ref{fig:envs}). The environments come with demonstrations which are gathered using virtual reality by a human. 

\paragraph{Robodesk} The Robodesk environment \citep{kannan2021robodesk} consists of a simulated Franka Emika Panda robot interacting with a table where multiple tasks are possible. The version of the simulated robot in the Robodesk environment only includes 5 DoFs (vs the 7 DoFs available, 2 were made not controllable). We evaluate AQuaPlay on the 9 base tasks described in Robodesk: \texttt{open\_slide}, \texttt{open\_drawer}, \texttt{push\_green}, \texttt{stack}, \texttt{upright\_block\_off\_table}, \texttt{flat\_block\_in\_bin}, \texttt{flat\_block\_in\_shelf}, \texttt{lift\_upright\_block}, \texttt{lift\_ball}.

We used the RLDS creator \url{github.com/google-research/rlds-creator} to generate play data, together with a Nintendo Switch Pro Controller. The data is composed by 50 episodes of approximately 3 minutes where the goal of the demonstrator is to \textit{interact} with the different elements of the environment.

\subsection{Hyperparameter Selection Procedure}
In the section we provide the hyperparameter selection procedure for the different setups. For the RLfD setting (Section~\ref{sec:lfd}) and the IL setting (Section~\ref{sec:il}) the number of hyperparameters is prohibitive to perform grid search. Therefore, we propose to sample hyperparameters uniformly within the set of all possible hyperparameters. For each environment, we sample 1000 configurations of hyperparameters, and train each algorithm including the baselines. We compute the average success rate of each individual value on the top 50\% of all corresponding configurations (since poorly performing configurations are less informative) and select the best performing hyperparameter value independently. This procedure enables to \textbf{1)} limit combinatorial explosion with the number of hyperparameters \textbf{2)} provide a fair evaluation between the baselines and the proposed algorithms as they all rely on the same amount of compute. In the \ifthenelse{\boolean{arxiv}}{the website}{supplementary material}, we provide histograms detailing the influence of each hyperparameter. For the RLfP setting, we fixed the parameters related to the DQN algorithm with the ones selected in the RLfD setting to limit the hyperparameter search, which enables to perform grid search for 3 seeds, and select the best set of hyperparameters.

\subsection{Reinforcement Learning with demonstrations}

\subsubsection{AQuaDQN \label{sec:lfd-aquadqn}}
We learn the discretization of the action space using the \aquadem{} framework. The architecture of the network is a common hidden layer of size $256$ with relu activation, and a subsequent hidden layer of size $256$ with relu activation for each action. We minimize the \aquadem{} loss using the Adam optimizer and dropout regularization.

We train a DQN agent on top of the discretization learned by the \aquadem{} framework. The architecture of the Q-network we use is the default LayerNorm architecture from the Q-network of the ACME library \citep{hoffman2020acme}, which consists in a hidden layer of size 512 with layer normalization and tanh activation, followed by two hidden layers of sizes 512 and 256 with elu activation. We explored multiple $Q$-value losses for which we used the Adam optimizer: regular DQN \citep{mnih2015human}, double DQN with experience replay \citep{van2016deep,schaul2015prioritized}, and Munchausen DQN \citep{vieillard2020munchausen}; the latter led to the best performance. We maintain a fixed ratio of demonstration episodes and agent episodes in the replay buffer similarly to \citet{hester2018deep}. We also provide as a hyperparameter an optional minimum reward to the transitions of the expert to have a denser reward signal. The hyperparameter sweep for AQuaDQN can be found in Table~\ref{table:sweep-aquadqn}. The complete breakdown of the influence of each hyperparameter is provided in \ifthenelse{\boolean{arxiv}}{the website.}{ \texttt{hps\_lfd\_aquadqn.html} in the supplementary.}

\begin{table}[h]
  \centering
  \begin{tabular}{llr}
    \toprule
    Hyperparameter & Possible values \\
    \midrule
    aquadem learning rate & 3e-5, 0.0001, \textbf{0.0003}, 0.001, 0.003\\
    aquadem input dropout rate & 0, \textbf{0.1}, 0.3 \\
    aquadem hidden dropout rate & 0, \textbf{0.1}, 0.3 \\
    aquadem temperature & 0.0001, \textbf{0.001}, 0.01\\
    aquadem \# actions & \textbf{10}, 15, 20 \\
    \midrule
    dqn learning rate & 0.00003, \textbf{0.0001}, 0.003\\
    dqn n step & 1, \textbf{3}, 5\\
    dqn epsilon & 0.001, 0.01, \textbf{0.1}\\
    dqn ratio of demonstrations & 0, 0.1, \textbf{0.25}, 0.5 \\
    dqn min reward of demonstrations & None, \textbf{0.01} \\
    \bottomrule
  \end{tabular}
  \caption{Hyperparameter sweep for the AQuaDQN agent}
  \label{table:sweep-aquadqn}
\end{table}

When selecting hyperparameters specifically for Relocate, for Figure~\ref{fig:lfd_relocate}, the main difference in the chosen values is a dropout rate set to 0.

\subsubsection{SAC and SACfD \label{app:lfd_sac}}
We reproduced the authors' implementations (with an adaptive temperature) and use MLP networks for both the actor and the critic with two hidden layers of size $256$ with relu activation. We use an Adam optimizer to train the SAC losses. We use a replay buffer of size $1$M, and sample batches of size $256$. We introduce a parameter of gradient updates frequency $n$ which indicates a number of $n$ gradient updates on the SAC losses every $n$ environment steps. SACfD is a version of SAC inspired by DDPGfD \citep{vecerik2017leveraging} where we add expert demonstrations to the replay buffer of the SAC agent with a ratio between the agent episodes and the demonstration episodes which is a hyperparameter. We also provide as a hyperparameter an optional minimum reward to the transitions of the expert to have a denser reward signal. We found that the best hyperparameters for SAC are the same for SACfD. The HP sweep for SAC and SACfD can be found in Table~\ref{table:sweep-sac} and Table~\ref{table:sweep-sacfd}. The complete breakdown of the influence of each hyperparameter is provided in \ifthenelse{\boolean{arxiv}}{the website.}{\texttt{hps\_lfd\_sac.html} and \texttt{hps\_lfd\_sacfd.html} in the supplementary.}

\begin{table}[h]
  \centering
  \begin{tabular}{llr}
    \toprule
    Hyperparameter & Possible values \\
    \midrule
    learning rate & 3e-5, \textbf{1e-4}, 0.0003 \\
    n step & 1, 3, \textbf{5} \\
    tau & \textbf{0.005}, 0.01, 0.05 \\
    reward scale & 0.1, 0.3, \textbf{0.5} \\
    \bottomrule
  \end{tabular}
  \caption{Hyperparameter sweep for SAC.}
  \label{table:sweep-sac}
\end{table}

\begin{table}[h]
  \centering
  \begin{tabular}{llr}
    \toprule
    Hyperparameter & Possible values \\
    \midrule
    learning rate & 3e-5, \textbf{1e-4}, 0.0003 \\
    n step & 1, 3, \textbf{5} \\
    tau & \textbf{0.005}, 0.01, 0.05 \\
    reward scale & 0.1, \textbf{0.3}, 0.5 \\
    \midrule
    ratio of demonstrations & 0, \textbf{0.001}, 0.1, 0.25 \\
    mini reward of demonstrations & None, \textbf{0.01}, \textbf{0.1}\\ 
    \bottomrule
  \end{tabular}
  \caption{Hyperparameter sweep for SACfD.}
  \label{table:sweep-sacfd}
\end{table}

\subsection{Imitation Learning}

\subsubsection{AQuaGAIL \label{app:aquagail}}
We learn the discretization of the action space using the AQuaDem framework. The architecture ofthe network is a common hidden layer of size 256 with relu activation, and a subsequent hidden layer of size 256 with relu activation for each action. We minimize the AQuaDem loss using the Adam optimizer and dropout regularization. The discriminator is a MLP whose number of layers, number of units per layers are hyperparameters. We use the Adam optimizer with two possible regularization scheme: dropout and weight decay. The discriminator outputs a value $p$ from which we compute three possible rewards $-\log(p)$, $-0.5\log(p) + \log(1-p)$, $\log(1-p)$ corresponding to the reward balance hyperparameter. The direct RL algorithm is Munchausen DQN, with the same architecture and hyperparameters described in Section~\ref{sec:lfd-aquadqn}. The hyperparameter sweep for AQuaGAIL can be found in Table~\ref{table:sweep-aquagail}. The complete breakdown of the influence of each hyperparameter is provided in \ifthenelse{\boolean{arxiv}}{the website}{\texttt{hps\_il\_aquagail.html} in the supplementary.}

\begin{table}[h!]
  \centering
  \begin{tabular}{llr}
    \toprule
    Hyperparameter & Possible values \\
    \midrule
    discriminator learning rate & 1e-7, 3e-7, \textbf{1e-6}, 3e-5, 1e-4\\
    discriminator num layers & \textbf{1}, 2 \\
    discriminator num units & 16, \textbf{64}, 256 \\
    discriminator regularization & none, \textbf{dropout}, weight~decay \\
    discriminator weight decay & 5, 10, 20\\
    discriminator input dropout rate & \textbf{0.5}, 0.75\\
    discriminator hidden dropout rate & \textbf{0.5}, 0.75\\
    discriminator observation normalization & \textbf{True}, False\\
    discriminator reward balance & 0., \textbf{0.5}, 1.\\
    \midrule
    dqn learning rate & \textbf{3e-5}, 1e-4, 3e-4\\
    dqn n step & \textbf{1}, 3, 5 \\
    dqn epsilon & 0.001, \textbf{0.01}, 0.1 \\
    \midrule
    aquadem learning rate &  3e-5, 1e-4, \textbf{3e-4}, 1e-3, 3e-3\\
    aquadem temperature & 0.0001, \textbf{0.001}, 0.01 \\
    aquadem num actions & \textbf{10}, 15, 20 \\
    aquadem input dropout rate & \textbf{0}, 0.1, 0.3 \\ 
    aquadem hidden dropout rate & \textbf{0}, 0.1, 0.3 \\     
    \bottomrule
  \end{tabular}
  \caption{Hyperparameter sweep for the AQuaGAIL agent.}
  \label{table:sweep-aquagail}
\end{table}

\subsubsection{GAIL}
We used the same discriminator architecture and hyperparameters as the one described in Section~\ref{app:aquagail}. The direct RL agent is the SAC algorithm whose architecture and hyperparameters are described in Section~\ref{app:lfd_sac}. The hyperparameter sweep for GAIL can be found in Table~\ref{table:sweep-gail}. The complete breakdown of the influence of each hyperparameter is provided in \ifthenelse{\boolean{arxiv}}{the website.}{\texttt{hps\_il\_gail.html} in the supplementary.} 

\begin{table}[h!]
  \centering
  \begin{tabular}{llr}
    \toprule
    Hyperparameter & Possible values \\
    \midrule
    discriminator learning rate & 1e-7, \textbf{3e-7}, 1e-6, 3e-5, 1e-4\\
    discriminator num layers & \textbf{1}, 2 \\
    discriminator num units & 16, 64, \textbf{256} \\
    discriminator regularization & none, dropout, \textbf{weight decay} \\
    discriminator weight decay & 5, \textbf{10}, 20\\
    discriminator input dropout rate & 0.5, 0.75\\
    discriminator hidden dropout rate & 0.5, 0.75\\
    discriminator observation normalization & \textbf{True}, False\\
    discriminator reward balance & 0., \textbf{0.5}, 1.\\
    \midrule
    sac learning rate & 3e-5, \textbf{1e-4}, 3e-4\\
    sac n step & 1, 3, \textbf{5} \\
    sac tau & 0.005, 0.01, \textbf{0.05} \\
    sac reward scale & 0.1, 0.3, \textbf{0.5}\\
    \bottomrule
  \end{tabular}
  \caption{Hyperparameter sweep for the discriminator part of the GAIL agent.}
  \label{table:sweep-gail}
\end{table}

\subsubsection{Behavioral Cloning}

The BC network is a MLP whose number of layers, number of units per layers and activation functions are hyperparameters. We use the Adam optimizer with two possible regularization scheme: dropout and weight decay. The observation normalization hyperparameter is set to True when each dimension of the observation are centered with the mean and standard deviation of the observations in the demonstration dataset. The complete breakdown of the influence of each hyperparameter is provided in \ifthenelse{\boolean{arxiv}}{the website.}{ \texttt{hps\_il\_bc.html} in the supplementary.} 

\begin{table}[h!]
  \centering
  \begin{tabular}{llr}
    \toprule
    Hyperparameter & Possible values \\
    \midrule
    learning rate & 1e-5, 3e-5, 1e-4, \textbf{3e-4}, 1e-3\\
    num layers & \textbf{1}, 2. 3 \\
    num units & 16, 64, \textbf{256} \\
    activation & relu, \textbf{tanh} \\
    observation normalization & \textbf{True}, False \\
    weight decay & \textbf{0}, .01, 0.1\\
    input dropout rate & \textbf{0}, 0.15, 0.3\\
    hidden dropout rate & \textbf{0}, 0.25, 0.5\\
    \bottomrule
  \end{tabular}
  \caption{Hyperparameter sweep for the BC agent.}
  \label{table:sweep-bc}
\end{table}

\subsubsection{Mixture Density Networks}

The MDN network is trained similarly as the action candidates network from the \aquadem{} framework. The architecture of the network is a common hidden layer of size $256$ with relu activation, and a subsequent hidden layer of size $256$ with relu activation for each action. We add another head from the common hidden layer that represents the \textit{logits} $p_i$ of each action candidate $a_i$. We train the MDN network with a modified version of the \aquadem{} loss using the Adam optimizer and dropout regularization:

\begin{equation}
\left\{
\begin{aligned}
&\min_{\Psi} \E_{s, a \sim \mathcal{D}} \big[ - T \log \big( \sum^K_{k=1} \exp( \frac{-\|\Psi_k(s) - a \|^2}{T}) \big) \big], \\ 
&\min_{p} \E_{s, a \sim \mathcal{D}} \big[ - \sum^K_{k=1} \frac{\exp(p_k)}{\sum_{k'} \exp(p_{k'})} \exp( \frac{-\|\Psi_k(s) - a \|^2}{T}) \big].
\end{aligned}
\right.
\end{equation}

At inference, we sample actions with respect to the logits. 

\begin{table}[h!]
  \centering
  \begin{tabular}{llr}
    \toprule
    Hyperparameter & Possible values \\
    \midrule
    learning rate & 3e-5, 0.0001, \textbf{0.0003}, 0.001, 0.003\\
    input dropout rate & \textbf{0}, 0.1\\
    hidden dropout rate & \textbf{0}, 0.1 \\
    temperature & 0.0001, \textbf{0.001}\\
    \# actions & 5, \textbf{10}, 20 \\
    \bottomrule
  \end{tabular}
  \caption{Hyperparameter sweep for the MDN agent.}
  \label{table:sweep-bc}
\end{table}

\subsection{Reinforcement Learning with play data}

\subsubsection{SAC}
We used the exact same implementation as the one described in Section~\ref{app:lfd_sac}. The HP sweep can be found in Table~\ref{table:sweep-play-sac}.

\begin{table}[h!]
  \centering
  \begin{tabular}{llr}
    \toprule
    Hyperparameter & Possible values \\
    \midrule
    learning rate & 1e-5, 3e-5, 3e-4, \textbf{1e-4} \\
    n step & 1, 3, \textbf{5}\\
    reward scale & 0.1, \textbf{1}, 10 \\
    tau & 0.005, \textbf{0.01}, 0.05\\
    \bottomrule
  \end{tabular}
  \caption{Hyperparameter sweep for SAC for the Robodesk environment. The best hyperparameter set was chosen as the one that maximizes the performance on average on all tasks.}
  \label{table:sweep-play-sac}
\end{table}

\subsubsection{DQN with Naive Discretization}
We used the exact same implementation as the one described in Section~\ref{sec:lfd-aquadqn}, and also use the best hyperparameters found in the RLfD setting. As the action space is $(-1, 1)^5$, we use three different discretization meshes: $\{-1, 1\}$, $\{-1, 0, 1\}, \{-1, -0.5, 0., 0.5, 1\}$ which induce a discrete action space of dimension $2^5, 3^5, 5^5$ respectively. We refer to the resulting algorithm as BB-2, BB-3, and BB-5 (where BB stands for 
``Bang-bang'').

\subsubsection{AQuaPlay}
We used the exact same implementation as the one described in Section~\ref{sec:lfd-aquadqn}, and also use the best hyperparameters found in the RLfD setting for the Munchausen DQN agent. We performed a sweep on the discretization step that we report in Table~\ref{table:sweep-aquaplay}.

\begin{table}[h!]
  \centering
  \begin{tabular}{llr}
    \toprule
    Hyperparameter & Possible values \\
    \midrule
    learning rate & 0.0001, 0.0003, \textbf{0.001}\\
    dropout rate & 0, \textbf{0.1}, 0.3 \\
    temperature & \textbf{1e-4}, 1e-3, 1e-2\\
    \# actions &  10, 20, \textbf{30}, 40\\
    \bottomrule
  \end{tabular}
  \caption{Hyperparameter sweep for the AQuaPlay agent. The best hyperparameter set was chosen as the one that maximizes the performance on average on all tasks.}
  \label{table:sweep-aquaplay}
\end{table}

\end{document}